\documentclass[conference]{IEEEtran}
\IEEEoverridecommandlockouts
\usepackage{cite}
\usepackage{amsmath,amssymb,amsfonts}
\usepackage{algorithmic}
\usepackage{graphicx}
\usepackage{textcomp}
\usepackage{xcolor}
\def\BibTeX{{\rm B\kern-.05em{\sc i\kern-.025em b}\kern-.08em
    T\kern-.1667em\lower.7ex\hbox{E}\kern-.125emX}}
\begin{document}

\title{A Hybrid Model-Based and Model-Free Framework
for Active Multi-View Viewpoint Optimization
in Sonar Target Recognition\\
}

\author{\IEEEauthorblockN{
Yongkyoon Park}
\IEEEauthorblockA{
\textit{University of Florida}\\
Gainesville, FL, United States of America \\
yo.park@ufl.edu}
\and
\IEEEauthorblockN{
Jane Shin}
\IEEEauthorblockA{
\textit{University of Florida}\\
Gainesville, FL, United States of America \\
jane.shin@ufl.edu}
}

\maketitle

\begin{abstract}
This paper presents a hybrid model-based and model-free framework for active multi-view target recognition using forward-looking sonar. A convolutional neural network (CNN) provides data-driven observation likelihoods, while Radon-based orientation estimation enables viewpoint-aware sensing without requiring angle annotations. During training, an information-gain–based reward guides a Proximal Policy Optimization (PPO) agent to learn a belief-aware viewpoint selection policy offline. At deployment, the learned policy performs real-time viewpoint selection using only CNN-based belief updates, eliminating the need for computationally expensive online POMDP tree search. Experiments on a marine-debris forward-looking sonar dataset demonstrate that the proposed approach achieves competitive recognition accuracy while reducing sensing steps and motion cost compared to model-based baselines.
\end{abstract}


\section{Introduction}
Unmanned underwater vehicles (UUVs) equipped with forward-looking sonar (FLS) rely on sequential observations from multiple viewpoints to achieve reliable target recognition. However, sonar imagery is inherently challenging due to severe noise, clutter, and strong viewpoint-dependent appearance variations. As a result, effective viewpoint selection becomes a critical component for active sonar perception.

Existing multi-view planning approaches typically rely on analytic information-theoretic criteria or model-based Partially Observable Markov Decision Process (POMDP) solvers to select informative viewpoints. Although these methods provide principled decision-making frameworks, they often suffer from high computational complexity or limited adaptability when applied to real sonar imagery.

To address these challenges, this work investigates a hybrid framework that combines belief-based reasoning and information-theoretic guidance with model-free reinforcement learning to enable efficient and real-time viewpoint selection.

\section{Related Work}

\subsection{Information-theoretic method}

Information-theoretic approaches \cite{b1, b8} have been widely used for active perception and viewpoint planning. These methods select sensing actions that maximize expected information gain (EIG) or reduce entropy, representing the belief uncertainty. Such criteria provide a principled mechanism for selecting informative viewpoints and have been applied to various multi-view recognition and exploration problems. However, most information-theoretic methods rely on myopic decision rules, optimizing only the immediate information gain without considering long-term sensing outcomes. As a result, they may produce suboptimal sensing trajectories when sequential observations and future decisions must be considered.

\subsection{Model-based POMDP method}

Partially observable Markov Decision Processes (POMDP) provide a theoretically ideal and principled framework for sequential decision making under uncertainty and have been widely used for active sensing and multi-view recognition \cite{b2, b3, b10}. In these approaches, the sensing agent maintains a belief distribution over possible target classes and selects actions that maximize the expected long-term reward through belief-space planning. Although this formulation enables theoretically optimal decision making, solving POMDP typically requires online tree search or value iteration during the execution, which introduces significant computational overhead. This limitation makes POMDP-based approaches difficult to deploy in real-time sensing systems, where decisions must be made quickly.

\subsection{Model-free Reinforcement Learning method}

Model-free reinforcement learning has recently been explored as an alternative approach for active perception \cite{b5}. These methods learn sensing policies directly from data, allowing fast decision making at deployment time without explicit online planning. However, purely model-free approaches often do not explicitly reason about belief uncertainty or information gain, which can lead to inefficient sensing behaviors. In addition, learning effective policies typically requires large amounts of training data and extensive training time.

\section{Problem Formulation}
We consider the problem of classifying a static underwater object using an Unmanned underwater vehicle (UUV) equipped with a forward-looking sonar (FLS). The AUV can acquire sonar images from a discrete set of viewpoints arranged along a circular trajectory around the object.

Due to severe noise and viewpoint-dependent appearance variations in sonar imagery, a single observation is often insufficient for reliable classification. Therefore, the UUV must actively select informative viewpoints to improve classification confidence while operating under limited motion and sensing budgets.

\subsection{POMDP Model}

The problem is formulated as a partially observable Markov decision process (POMDP)
$$\mathcal{P} = (X, A, Z, T, O, R, \gamma)$$
where \begin{itemize}
    \item $X$ is the set of possible object classes,
	\item $A$ is the set of viewpoint-selection actions including a terminal stop action,
	\item $Z$ denotes the observation space of sonar images,
	\item $T(x'|x,a)$ is the state transition model,
	\item $O(z|x,a)$ represents the observation likelihood,
	\item $R$ is the reward function, and
	\item $\gamma$ is the discount factor.
\end{itemize}

Because the target class remains constant during observation, the transition model is deterministic.

\subsection{Belief Representation}

Since the true object class x is unknown, the agent maintains a belief distribution over classes
$$b_t(x) = P(x \mid z_{1:t}, a_{1:t-1})$$
which represents the posterior probability of each class given the observation and action history.

In a general POMDP, the belief is updated according to
$$b_{t+1}(x') =
\eta \, O(z_t \mid x', a_t)
\sum_{x \in X} T(x' \mid x, a_t) \, b_t(x),$$
where $T(x'|x,a)$ is the state transition model, $O(z|x,a)$ is the observation likelihood, and $\eta$ is a normalization constant.

In the target recognition problem considered in this work, the hidden state corresponds to the object class, which remains constant during the observation process. Therefore, the transition model becomes deterministic.

Under this assumption, the belief updates simplifies to
$$b_{t+1}(x) = \eta \, O(z_t|x,a_t)\, b_t(x)$$
which corresponds to a Bayesian update using the observation likelihood.

\subsection{Action Space}

At each step, the agent selects an action $a_t \in A$ that corresponds to moving to one of $N$ candidate viewpoints around the target or issuing a stop action to terminate observation and perform classification.

Each movement incurs a distance-dependent motion cost.

\subsection{Observation Model}

Each sensing action produces a sonar observation $z_t \in Z$.
The observation likelihood
$$O(z_t \mid x, a_t)$$
represents the probability of observing sonar image $z_t$ when the true object class is $x$ and the agent executes action $a_t$.

This likelihood is used to update the belief over object classes according to the Bayesian update rule described in the previous section.

In practice, the observation likelihood is estimated using a data-driven observation model, which will be described in the proposed framework section.

\subsection{Reward Function}

The reward function is designed to encourage the agent to obtain informative observations while minimizing unnecessary sensing actions and motion cost. The reward at time step $t$ is defined as
$$R_t = R_{\text{class}} - \lambda_1 d_{\text{move}}(a_t) - \lambda_2 + R_{\text{info}},$$
where \begin{itemize}
    \item 	$R_{\text{class}}$ provides a reward when the agent correctly terminates the sensing process with a confident classification decision,
	\item $d_{\text{move}}(a_t)$ represents the motion cost associated with executing action $a_t$,
	\item $\lambda_1$ and $\lambda_2$ are weighting parameters that penalize excessive motion and unnecessary sensing steps,
	\item $R_{\text{info}}$ is an information-driven reward that encourages the agent to select viewpoints that reduce classification uncertainty.
\end{itemize}

The information-driven reward guides the agent to actively acquire observations that improve the belief over object classes. The specific formulation of this information-based reward is described in the proposed framework section.

\subsection{Objects}

The objective of the agent is to learn a policy $\pi(a \mid s)$ that selects sensing actions to maximize the expected cumulative reward:
$$\pi^* =
\arg\max_{\pi}
\mathbb{E}
\left[
\sum_{t=0}^{T}
\gamma^t R_t
\right].$$

In classical POMDP-based planning approaches, this policy is typically computed through online belief-space tree search, which can be computationally expensive. In contrast, this work learns the policy offline using reinforcement learning while preserving the belief update and information-theoretic structure of the problem.

\section{Proposed Framework}

This section presents the proposed hybrid framework for active multi-view sonar target recognition. The framework integrates belief-based reasoning with model-free policy learning to enable efficient viewpoint selection under partial observability.

The system consists of four key components: a CNN-based observation likelihood model, geometric orientation estimation using the Radon transform, a belief-aware state representation, and a PPO-based policy learning framework guided by an information-driven reward.

\begin{figure} [h]
    \centering
    \includegraphics[width=0.8\linewidth]{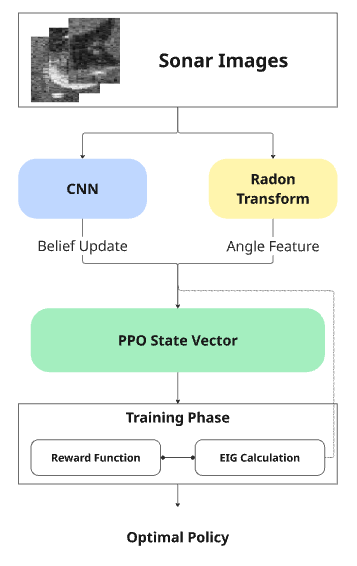}
    \caption{Proposed Framework Overview}
    \label{fig:placeholder}
\end{figure}

\subsection{CNN-Based Observation Likelihood}
To estimate the likelihood of each target class given a sonar observation, a convolutional neural network (CNN) classifier is used to process the incoming sonar images.

Let $o_t$ denote the sonar observation obtained at time step $t$. The CNN processes the observation and outputs a class probability vector
$$p_\theta(x \mid o_t)$$
which represents the posterior probability of each target class.

In a classical POMDP formulation, belief updates require the generative observation likelihood $P(o_t \mid x)$. However, constructing a generative sonar observation model is challenging due to the complex acoustic scattering, shadowing, and noise characteristics present in real sonar imagery.

To address this limitation, the proposed framework uses the CNN discriminative posterior as a surrogate observation likelihood. Assuming a uniform class prior, Bayes’ rule implies
$$P(o_t \mid x) \propto P(x \mid o_t)$$
which allows the CNN output to serve as an approximation of the observation likelihood.

These likelihood estimates are used to update the belief distribution over target classes, enabling the system to progressively refine its class estimate as new observations are collected.

\subsection{Radon-Based Orientation Estimation}

In addition to classification likelihood, the framework extracts geometric information from the sonar images to characterize the relative aspect sensing configuration.

The Radon transform \cite{b11} is applied to the sonar image to estimate the dominant structural orientation of the observed target. By analyzing projection profiles across multiple angles, the Radon transform identifies the orientation that maximizes structural alignment in the image.

Let $\theta_t$ denote the estimated orientation at time step $t$. Rather than using the absolute orientation directly, the framework computes the relative aspect angle
$$\Delta\theta_t = \theta_t - \theta_{t-1}$$
which represents the change in viewing direction between consecutive observations.

The relative aspect angle provides a geometric cue that indicates how the observation geometry evolves over time. In sonar imaging, target appearance varies significantly with viewing direction due to acoustic scattering and shadowing effects. As a result, observations obtained from sufficiently different aspect angles are more likely to reveal complementary structural information about the target.

By incorporating the relative aspect angle into the policy state representation, the agent can reason about viewpoint transitions in a geometry-aware manner and avoid uninformative observations.

\subsection{Policy State Representation}

To support belief-aware decision making, the policy state representation integrates information about classification uncertainty, sensing geometry, and viewpoint coverage.

The policy input state vector includes
\begin{itemize}
    \item Belief vector $b_t$, representing the posterior probability of each target class,
	\item One-hot encoding of the current viewpoint,
	\item Binary mask indicating previously visited viewpoints,
	\item Relative orientation features ($\cos\Delta\theta_t$, $\sin\Delta\theta_t$).
\end{itemize}

This representation allows the policy to reason about both semantic uncertainty and sensing geometry when selecting the next viewpoint.

\subsection{Information-Guided Reward Function}

To encourage the agent to actively select informative viewpoints, the reward function incorporates an information-driven component based on Expected Information Gain which means belief entropy reduction.

The expected information gain is defined as
$$EIG_t = H(b_t) - H(b_{t+1}),$$
where $H(b)$ denotes the entropy of the belief distribution.

A positive information gain indicates that the selected observation reduces classification uncertainty. By rewarding actions that maximize information gain, the policy is encouraged to explore viewpoints that provide new discriminative evidence about the target.

The overall reward combines the information gain term with penalties for motion cost and unnecessary sensing steps, as described in the problem formulation.

\subsection{PPO-Based Policy Learning}

The viewpoint selection policy is learned using Proximal Policy Optimization (PPO) \cite{b6}, a model-free reinforcement learning algorithm known for stable and efficient policy optimization.

During training, the agent interacts with the simulated sensing environment and collects trajectories consisting of states, actions, observations, and rewards. The PPO algorithm updates the policy parameters using these trajectories to maximize the expected cumulative reward.

After training is completed, the learned policy is deployed for real-time viewpoint selection. At each sensing step, the system processes the incoming sonar observation, updates the belief state, computes the relative orientation feature, and constructs the policy state vector.

The trained policy then selects the next sensing action through a single forward pass of the policy network. Unlike classical POMDP solvers that require computationally expensive online tree search, the learned policy enables fast decision making during execution while preserving belief-aware reasoning learned during training.

\section{Experiments and Results}

This section evaluates the effectiveness of the proposed framework for active multi-view sonar target recognition. The experiments aim to answer two main questions. First, we examine whether the proposed method improves recognition performance through informative viewpoint selection. Second, we analyze whether the learned policy enables efficient real-time decision making compared to classical planning-based approaches.

\subsection{Dataset and Experimental Setup}

The proposed framework is evaluated using the marine debris forward-looking sonar dataset. The dataset contains sonar images of underwater objects collected using a forward-looking sonar (FLS) sensor. Each object instance is captured under multiple viewing angles using a controlled turntable setup, providing a set of observations that simulate multi-view sensing conditions.

To emulate the sensing process of an unmanned underwater vehicle (UUV), the sonar images are treated as observations obtained from discrete viewpoints arranged around the target object. In this study, the sensing viewpoints are discretized into eight equally spaced positions along a circular trajectory surrounding the target. At each sensing step, the agent selects an action that moves the sensor to one of the available viewpoints or terminates the sensing process.

Although the true sensing geometry in underwater environments is inherently three-dimensional, this study assumes that both the sensing platform and the target object lie on the same plane. Under this assumption, viewpoint transitions correspond to changes in the relative azimuth angle between the sensor and the target.

To better approximate real underwater sensing conditions, synthetic perturbations are introduced during training and evaluation. These include random occlusion and image jitter, which simulate partial visibility and sensor instability commonly encountered in practical sonar imaging scenarios.

\subsection{Baseline Methods}

To evaluate the proposed framework, we compare it with two representative approaches for active perception.

\subsubsection{\textbf{Model-based planning}}

The first baseline is a model-based planning approach \cite{b2, b3, b10} that performs viewpoint selection using belief-based planning. This method explicitly models the POMDP structure and selects actions by maximizing cumulative reward through online decision making.

Although model-based methods can explicitly reason about uncertainty and information gain, they typically require computationally expensive online planning procedures, such as a belief-space tree search.

\subsubsection{\textbf{Model-free reinforcement learning}}

The second baseline is a purely model-free reinforcement learning approach \cite{b5, b6} that learns a viewpoint selection policy directly from the data. In this method, the agent learns a policy using reinforcement learning without explicitly incorporating belief updates or structured uncertainty modeling.

Although model-free RL methods can learn policies efficiently through experience, they often struggle to reason about classification uncertainty because they lack an explicit belief representation.

\subsection{Recognition Performance}

We first evaluate the recognition capability of the proposed framework using the marine debris sonar dataset. 

\begin{table}[htbp]
\caption{Recognition Performance}
\begin{center}
\begin{tabular}{|c|c|c|c|}
\hline
\textbf{Table1}&\multicolumn{3}{|c|}{\textbf{Models}} \\
\cline{2-4} 
\textbf{Metrics} & \textbf{\textit{POMDP}}& \textbf{\textit{RL}}& \textbf{\textit{Proposed}} \\
\hline
Final Accuracy& 0.934 & 0.974 & 0.990 \\
Step to 90\% ACC & 1.63 & 0.40 & 0.34 \\
\hline
\end{tabular}
\label{tab1}
\end{center}
\end{table}

Table 1 summarizes the final classification accuracy and the sensing efficiency of each method. The sensing efficiency is measured using the average number of sensing steps required to reach 90\% recognition accuracy.

The results show that the proposed hybrid framework achieves the highest final recognition accuracy of 0.990, outperforming both the model-based POMDP baseline and the purely model-free reinforcement learning policy. The POMDP method achieves a final accuracy of 0.934, while the model-free PPO baseline reaches 0.974.

In addition to the final classification accuracy, Table 1 also reports the average number of sensing steps required to reach 90\% accuracy. This metric reflects how efficiently each method acquires informative observations during the active sensing process. The proposed method reaches 90\% accuracy at an average of 0.34 sensing steps, which is comparable to the model-free PPO policy (0.40) and significantly faster than the model-based POMDP baseline (1.63).

These results indicate that the proposed hybrid framework effectively balances recognition accuracy and sensing efficiency. By integrating belief-based reasoning with reinforcement learning, the proposed method is able to select informative viewpoints early in the sensing process while maintaining superior final classification performance.

\subsection{Computational Performance}

We next evaluate the computational efficiency of the proposed framework and compare it with the model-based and model-free baselines.

\begin{table}[htbp]
\caption{Computational Performance}
\begin{center}
\begin{tabular}{|c|c|c|c|}
\hline
\textbf{}&\multicolumn{3}{|c|}{\textbf{Models}} \\
\cline{2-4} 
\textbf{Metrics} & \textbf{\textit{POMDP}}& \textbf{\textit{RL}}& \textbf{\textit{Proposed}} \\
\hline
Decision Time per Step [ms] & 70.267 & 0.717 & 0.638 \\
Total Episode Time [s] & 0.2373 & 0.0043 & 0.0052 \\
\hline
\end{tabular}
\label{tab2}
\end{center}
\end{table}

Table 2 summarizes the average decision time per sensing step and the total episode execution time for each method. The model-based POMDP approach requires significantly higher computation time due to the online planning process. Specifically, the POMDP method requires 70.267 ms per decision step, resulting in a total episode time of 0.2373 s.

In contrast, both the model-free reinforcement learning policy and the proposed hybrid framework require only a single forward pass of the policy network to determine the next action. As a result, the decision time per step is reduced to 0.717 ms for the model-free policy and 0.638 ms for the proposed method.

The total episode execution time further highlights this difference. While the POMDP approach requires 0.2373 s, the model-free and proposed methods require only 0.0043 s and 0.0052 s, respectively. These results demonstrate that the proposed framework achieves computational efficiency comparable to model-free reinforcement learning while significantly outperforming model-based planning approaches.

This property is particularly important for unmanned underwater systems, where sensing decisions must be made under strict computational and time constraints.

\subsection{Trajectory Sample}

\begin{figure} [h]
        \centering
    \includegraphics[width=1\linewidth]{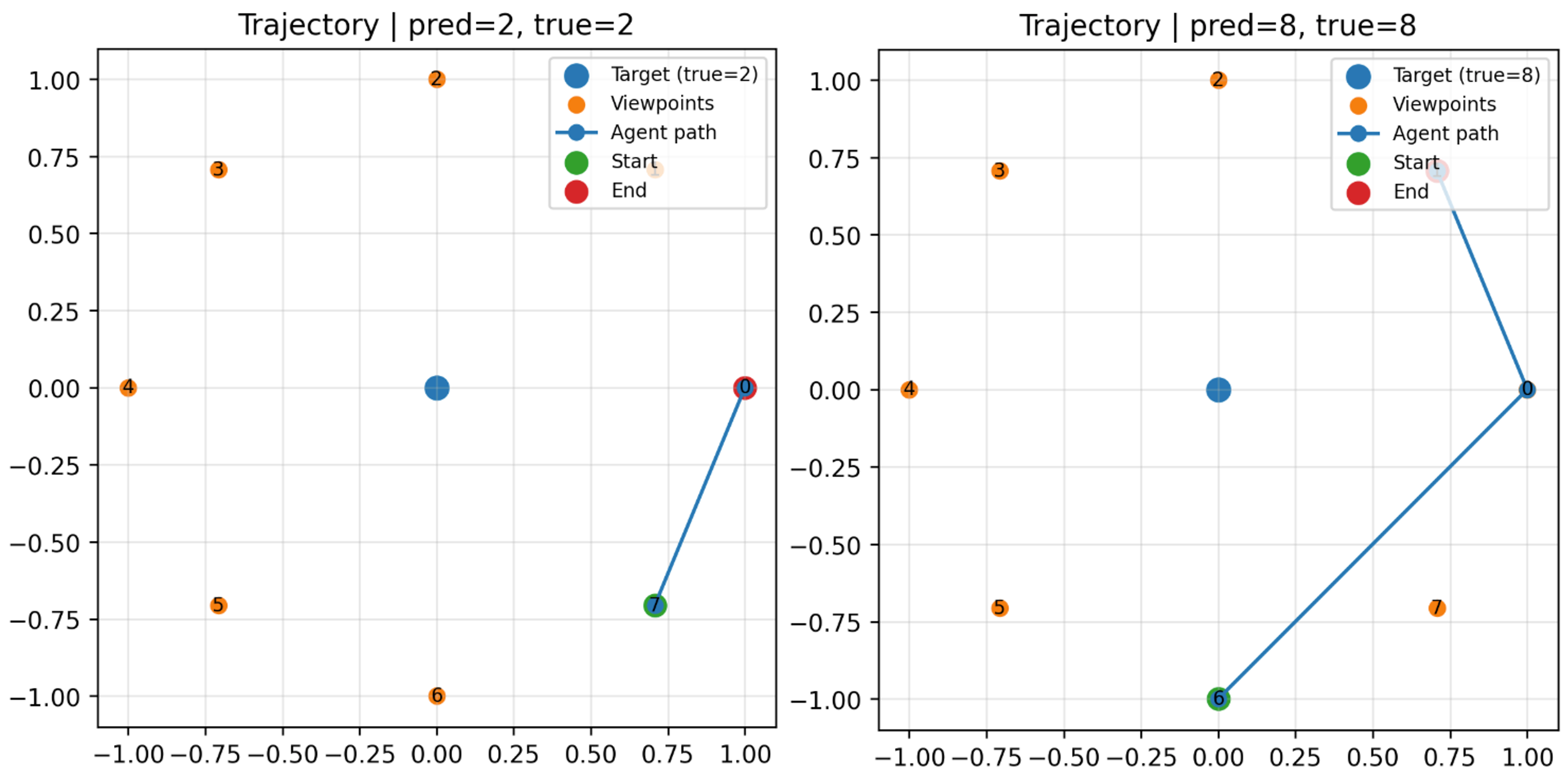}
    \caption{Trajectory Samples}
    \label{fig:placeholder}
\end{figure}

Fig. 2 illustrates representative sensing trajectories generated by the proposed policy during active viewpoint selection. Each trajectory shows the sequence of viewpoints selected by the agent while observing the target object from discrete positions arranged around the object.

In the left example, the agent requires only a small number (2-steps) of observations before terminating the sensing process. After acquiring an additional viewpoint, the agent reaches sufficient classification confidence and stops early. This demonstrates that the policy can quickly identify informative viewpoints and avoid unnecessary sensing actions when the object can be recognized with high certainty.

In contrast, the right example illustrates a more ambiguous case where the initial observations are not sufficiently informative. In this scenario, the agent continues to select additional viewpoints to reduce the uncertainty of the belief state before making the final classification decision. The resulting trajectory shows that the agent moves toward viewpoints that provide complementary information about the object.

These qualitative examples indicate that the proposed policy adapts its sensing strategy depending on the uncertainty of the current belief state. When the classification confidence increases rapidly, the agent terminates the sensing process early. Otherwise, it continues to explore additional viewpoints to collect more informative observations.

\section{Conclusion}

This paper presented a hybrid policy learning framework for active multi-view target recognition using forward-looking sonar. The proposed method integrates belief-based reasoning from POMDP formulations with model-free reinforcement learning to enable efficient viewpoint selection under uncertainty. During training, an information-gain–based reward guides the policy to acquire informative observations, while a CNN-based observation model provides likelihood estimates for belief updates. After training, the learned policy allows real-time viewpoint selection without requiring computationally expensive online POMDP planning.

Experimental results on a marine debris sonar dataset demonstrate that the proposed approach achieves higher recognition accuracy compared to both model-based POMDP planning and model-free reinforcement learning baselines. Furthermore, the proposed method reaches high recognition confidence with fewer sensing steps while maintaining computational efficiency comparable to model-free policies. Qualitative trajectory analysis further shows that the learned policy adaptively adjusts its sensing strategy depending on the uncertainty of the current belief state.

These results suggest that combining belief updates with reinforcement learning provides an effective framework for active perception in sonar-based object recognition tasks.


\end{document}